\documentclass[10pt,twocolumn,letterpaper]{article}

\usepackage{wacv}              

\pdfoutput=1
\usepackage{graphicx}
\usepackage{amsmath}
\usepackage{amssymb}
\usepackage{mathtools}
\usepackage{amsthm}
\usepackage{booktabs,colortbl}
\usepackage{multirow}
\usepackage{centernot}
\usepackage{bbm}
\usepackage[inline]{enumitem}
\usepackage[numbers]{natbib}
\usepackage[accsupp]{axessibility} 

\usepackage{pgfplots}
\DeclareUnicodeCharacter{2212}{−}
\usepgfplotslibrary{groupplots,dateplot}
\usetikzlibrary{patterns,shapes.arrows}
\pgfplotsset{compat=newest}

\newcommand{\zh}{Z}
\newcommand{\sh}{\hat{S}}
\newcommand{\bh}{\hat{B}}
\newcommand{\s}{L}
\newcommand{\laece}{\operatorname{LaECE}}
\newcommand{\dece}{\operatorname{D-ECE}}
\newcommand{\iou}{\operatorname{IoU}}
\newcommand{\kde}{\operatorname{\widehat{\operatorname{CE}}}}

\theoremstyle{plain}
\newtheorem{theorem}{Theorem}[section]
\newtheorem{proposition}[theorem]{Proposition}

\definecolor{maroon}{cmyk}{0,0.87,0.68,0.32}

\usepackage[pagebackref,breaklinks,colorlinks]{hyperref}

\usepackage[capitalize]{cleveref}
\crefname{section}{Sec.}{Secs.}
\Crefname{section}{Section}{Sections}
\Crefname{table}{Table}{Tables}
\crefname{table}{Tab.}{Tabs.}

\def\wacvPaperID{*****} 
\def\confName{WACV}
\def\confYear{2024}

\begin{document}

\title{Beyond Classification: Definition and Density-based Estimation of Calibration in Object Detection}

\author{Teodora Popordanoska\\
ESAT-PSI, KU Leuven \\
{\tt\small teodora.popordanoska@kuleuven.be}
\and
Aleksei Tiulpin  \\
HST Research Unit, University of Oulu  \\
{\tt\small aleksei.tiulpin@oulu.fi}
\and
Matthew B. Blaschko \\
ESAT-PSI, KU Leuven \\
{\tt\small matthew.blaschko@esat.kuleuven.be}
}
\maketitle

\begin{abstract}

Despite their impressive predictive performance in various computer vision tasks, deep neural networks (DNNs) tend to make overly confident predictions, which hinders their widespread use in safety-critical applications. 
While there have been recent attempts to calibrate DNNs, most of these efforts have primarily been focused on classification tasks, thus neglecting DNN-based object detectors. 
Although several recent works addressed calibration for object detection and proposed differentiable penalties, none of them are consistent estimators of established concepts in calibration. 
In this work, we tackle the challenge of defining and estimating calibration error specifically for this task. 
In particular, we adapt the definition of classification calibration error to handle the nuances associated with object detection, and predictions in structured output spaces more generally. Furthermore, we propose a consistent and differentiable estimator of the detection calibration error, utilizing kernel density estimation. Our experiments demonstrate the effectiveness of our estimator against competing train-time and post-hoc calibration methods, while maintaining similar detection performance. 
\end{abstract}

\section{Introduction}
\label{sec:intro}

Calibration is a property of a model, which directly translates to the ability to estimate its own predictive uncertainty, thereby facilitating safe and responsible deployment. Intuitively, a well-calibrated model produces confidence scores that accurately reflect the uncertainty associated with its predictions. In deep neural networks (DNNs), which are driving most of the current progress in machine learning and computer vision, this has become an emerging concern, as they can yield wrong predictions with high confidence \cite{guo2017calibration,ovadia2019can}. Unreliable uncertainty estimates produced by such models can lead to erroneous decision-making, which is especially risky in fields like autonomous driving~\cite{yurtsever2020survey} and medicine~\cite{dusenberry2020analyzing}.
The field of model calibration typically studies three sub-problems: \textit{estimating} calibration error, \textit{regularization} during training, and \textit{post-hoc} calibration. While in recent years there have been multiple advances in all of these domains, most of the existing literature studies calibration in the context of classification. Calibration in structured prediction problems, such as object detection (OD), have received substantially less attention, despite their increased integration in many safety-critical applications. While some attempts have been made to define, measure, enforce, and adjust calibration \cite{kueppers2020, munir2022towards, oksuz2023towards, munir2023bridging}, we argue that the field of OD lacks a solid mathematical foundation in both definition of calibration and estimation of calibration errors.

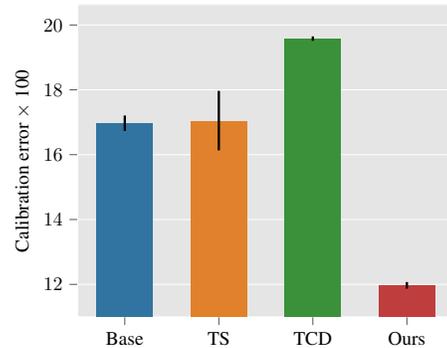
\begin{figure}[t]
    \centering
    \resizebox{0.35\textwidth}{!}{
\begin{tikzpicture}
\definecolor{color0}{rgb}{0.194607843137255,0.453431372549019,0.632843137254902}
\definecolor{color1}{rgb}{0.881862745098039,0.505392156862745,0.173039215686275}
\definecolor{color2}{rgb}{0.229411764705882,0.570588235294118,0.229411764705882}
\definecolor{color3}{rgb}{0.75343137254902,0.238725490196078,0.241666666666667}

\begin{axis}[
axis background/.style={fill=white!89.8039215686275!black},
axis line style={white},
scaled y ticks = false,
tick align=outside,
tick pos=left,
x grid style={white},
xmin=-0.5, xmax=3.5,
xtick style={color=white!33.3333333333333!black},
xtick={0,1,2,3},
xticklabels={Base,TS,TCD,Ours},
y grid style={white},
y tick label style={/pgf/number format/fixed},
ylabel={Calibration error \(\displaystyle \times\) 100},
ymajorgrids,
ymin=11, ymax=20.6302912112395,
ytick style={color=white!33.3333333333333!black}
]
\draw[draw=none,fill=color0,very thin] (axis cs:-0.3,0) rectangle (axis cs:0.3,16.97);
\draw[draw=none,fill=color1,very thin] (axis cs:0.7,0) rectangle (axis cs:1.3,17.05);
\draw[draw=none,fill=color2,very thin] (axis cs:1.7,0) rectangle (axis cs:2.3,19.58);
\draw[draw=none,fill=color3,very thin] (axis cs:2.7,0) rectangle (axis cs:3.3,11.97);
\path [draw=black, very thick]
(axis cs:0,16.7323626292015)
--(axis cs:0,17.2076373707984);

\path [draw=black, very thick]
(axis cs:1,16.1333987126346)
--(axis cs:1,17.9666012873654);

\path [draw=black, very thick]
(axis cs:2,19.5121036083433)
--(axis cs:2,19.6478963916567);

\path [draw=black, very thick]
(axis cs:3,11.8681554125149)
--(axis cs:3,12.0718445874851);

\end{axis}

\end{tikzpicture}
    }
    \caption{Calibration error of RetinaNet on Pascal VOC. The model without calibration (Base) is compared with post-hoc (temperature scaling; TS) and train-time calibration methods (TCD and Ours). Our KDE-based estimator effectively reduces the calibration error. The error bars represent 95\% CI. }
    \label{fig:enter-label}
\end{figure}

In spite of its importance, defining calibration for object detection is a challenging problem due to the variability of DNN-based detectors. In particular, there are ambiguities related to how many bounding boxes are returned, how to select a detection confidence threshold below which detections are rejected, what is the intersection over union (IoU) threshold for considering a ``correct`` detection, and so forth. There have been a few recent attempts to define calibration by either replacing accuracy with precision in the definition of calibration for classification~\cite{kueppers2020, pathiraja2023multiclass, munir2023bridging}, or by requiring that both classification and localization performances jointly match the confidence score~\cite{oksuz2023towards}. 

In this paper, we propose a general framework that unifies existing definitions of calibration in OD, and is flexible to parametrize the notion of a ``correct`` detection. Beyond defining calibration, we derive a \textit{consistent} and \textit{differentiable} estimator of calibration error in OD, relying on a kernel density estimator (KDE) recently proposed for classification~\cite{popordanoska2022}, and recommended among the best practices~\cite{maier2022metrics} for assessing calibration of classifiers. Due to its consistency and differentiability, our estimator of calibration error can be used not only as a reliable  tool to assess calibration, but also in calibration-regularized training of  popular detectors. We perform extensive experiments on MS COCO \cite{lin2014_coco}, Cityscapes \cite{cordts2016_cityscapes}, and PASCAL VOC \cite{everingham2012_pascalvoc}, and demonstrate that our estimator, when incorporated as an auxiliary loss during training, consistently reduces the calibration error across several object detectors. Furthermore, we show that finetuning with our loss is an effective way to improve calibration, while maintaining comparable average precision (AP), and without the need to re-train the models from scratch. 

In summary, our contributions are:
\begin{enumerate}[nosep]
    \item We propose a general unifying definition of calibration in OD, which addresses the nuances related to assessing the ``correctness`` of a detection.
    \item We develop a consistent and differentiable estimator of calibration error based on KDE~\cite{popordanoska2022} for OD.
    \item We perform rigorous empirical analysis of calibration of popular object detectors on several datasets.
    \item We demonstrate that our estimator can be used as an auxiliary train-time loss, which effectively reduces the calibration error, while maintaining similar average precision (AP). 
\end{enumerate}

\section{Related work}

\subsection{Calibration in classification}
Most of the prior work on developing techniques for improving calibration in the vision domain target the image classification task. 
Post-hoc methods \cite{guo2017calibration,Ma2021a,zadrozny2001,zadrozny2002} re-scale the output of a trained model using parameters learned on a validation set. The most successful method of this group is temperature scaling \cite{guo2017calibration}, where a single parameter $T$ is learned, usually by minimizing NLL, to scale the logits before applying the softmax function. 
Train-time calibration methods  \cite{popordanoska2022,kumar2018trainable,hebbalaguppe2022stitch,mukhoti2020, muller2020,liang2020} typically incorporate calibration-related auxiliary loss directly into the training process of a neural network. Even though these methods may introduce computational overhead, they do not require a hold-out validation set and have demonstrated superior performance in handling domain shift \cite{karandikar2021soft}, compared to post-hoc methods. 

\subsection{Calibration in object detection}
Recent works have shown that miscalibration is a concern not only in classification tasks, but also in the domain of object detection \cite{kueppers2020, pathiraja2023multiclass}. \citet{pathiraja2023multiclass} showed that existing calibration methods for classification are not as effective for calibrating object detectors. As a result, there have been a few efforts to establish the definition of calibration \cite{kueppers2020, oksuz2023towards} and to devise techniques to mitigate the issue. \citet{oksuz2023towards} propose a binning-based metric called Localisation-aware Expected
Calibration Error ($\laece$), and use it as part of existing post-hoc calibration approaches like linear regression, histogram binning \cite{zadrozny2001} and isotonic regression \cite{zadrozny2002}. 
In the realm of train-time calibration methods, several auxiliary loss terms have been proposed to be used in addition to the detection-specific loss. For instance, TCD \cite{munir2022towards} and MCCL \cite{pathiraja2023multiclass} were proposed to jointly calibrate the class-wise confidences and localization performance. 
Another recently introduced auxiliary loss is BPC \cite{munir2023bridging}, which is based on a heuristic that maximizes confidence scores for accurate predictions while minimizing scores for inaccurate predictions.
\citet{harakeh2021estimating} propose an energy score to train probabalistic detectors, which is empirically shown to improve calibration. 

However, none of the proposed binning-based metrics and trainable auxiliary losses fulfill both requirements of being consistent and differentiable estimators of calibration error in the context of object detection. In this paper, we derive a novel estimator for calibration error in object detection with those desired properties. 

\section{Methods}

\subsection{Defining calibration}
Several notions of different strength exist for multi-class calibration, including top-label, class-wise, and canonical calibration \cite{guo2017calibration,vaicenavicius2019,widmann2019}. However, following recent trends of training the classification branch of object detectors in a multi-label setting, i.e.\ by using $K$ independent binary classifiers \cite{lin2017_retinanet,tian2019_fcos, redmon2018yolov3}, we also favor the approach of defining calibration by considering $K$ binary object detectors, as the mutual-exclusion principle of multi-class classification does not hold (e.g.\ the area of an image occupied by a person may overlap with that of a bicycle). 

Intuitively, calibration requires that the confidence score of a prediction should be aligned with some notion of correctness. 
In classification tasks, a correct prediction is determined by the match between the predicted and ground truth labels. However, in object detection, assessing correctness is more nuanced, since it involves evaluating the degree of overlap between two sets of pixels (e.g. bounding boxes). To account for this, we will introduce a similarity measure (e.g. $\iou$) and a so-called link function, so that we can define a family of notions of calibration that also consider the degree of correctness of a given prediction. 
The formal definitions are given below.

\paragraph{Binary classification}
Let $X \in \mathcal{X}$ and $Y \in \mathcal{Y} =\{0, 1\}$ be random variables denoting the input and target, given by a joint distribution $P$.
Let $f$ be a neural network with $f(X) = \sh$, where $\sh \in [0,1]$ denotes the model's confidence that the label is 1. 
Then $f$ is said to satisfy \emph{binary calibration} \cite{brocker_2009reliability,zadrozny2002,kumar2019verified} if: 
\begin{equation}
    \label{eq:binary_calibration}
    \mathbb{P} \left(Y=1 \mid \sh = s\right)=s, \quad\forall s \in [0,1] .
\end{equation}

\paragraph{Object detection}
Given a set of images $\{X_i\}_{1\leq i \leq n}$, the goal of an object detector is to predict bounding boxes, class labels and confidence scores for the objects in $\{X_i\}$.
Let $g_k$ be a binary object detector for class $k$ with $g_k(X_i) = \{(\sh_{ij}, \bh_{ij})\}_{1\leq j \leq m_{ik}  }$, where $m_{ik}$ boxes are predicted for image $X_i$ and class $k$, $\bh_{ij} \in \mathcal{B} = \mathbb{R}^4$ denotes a predicted bounding box,\footnote{We note that in general this can be any set of pixels, not necessarily limited to the ones defined by a bounding box. We adopt this notation for convenience, but it does not change the mathematical principles involved.} and $\sh_{ij} \in [0,1]$ its corresponding score. 

Let $\s : \mathcal{B} \times \mathcal{B} \rightarrow [0,1]$ denote a similarity measure, for example $\iou$, Dice score, Hamming similarity etc.
Let $\psi: [0,1] \rightarrow [0,1]$ denote a monotonic function, which we will refer to as a \textit{link} function. Some choices for the link function in the context of object detectors may include an identity function $\psi(\s) = \s $, or a piece-wise linear ramp function parametrized with $\alpha\leq \beta \in [0, 1]$ as:
\begin{equation}
    \psi(\s) = \begin{cases} 
      0 & \s \leq \alpha \\
      \frac{\s-\alpha}{\beta-\alpha} & \alpha < \s < \beta \\
      1 & \s \geq \beta
   \end{cases} .
\end{equation}
Two special cases of this function that could be of interest are a step (threshold) function ($0<\alpha = \beta \leq 1)$, and a piece-wise linear (hinge) function ($\alpha = 0.5, \beta =1$).

Let $\zh := \psi(\s(\bh,B^{*}))$ be a random variable that denotes the ``correctness`` of a detection, and $B^{*}$ be the ground-truth box that $\bh$ matches with.
If $\psi$ is a threshold function then $\zh \in \{0, 1\}$ denotes whether the predicted bounding box matched a ground truth box with $\s \geq \beta$, and if $\psi$ is an identity function then $\zh \in [0, 1]$ represents the degree of ``correctness`` of the detection, e.g.\, the IoU between predicted and ground truth box. Then we define \textit{calibration for object detection} as:
\begin{equation}
    \label{eq:binary_calibration_od}
    \mathbb{P} \left(\zh \mid \sh = s\right) = s, \quad\forall s \in [0,1].
\end{equation}

The introduction of a link function $\psi$, provides a concise way to relate the notion of calibration to existing literature. For example, by letting $\psi$ be a threshold function and taking $\s = \iou$, we recover the definition of calibration used in \cite{kueppers2020,munir2022towards, munir2023bridging,pathiraja2023multiclass}, i.e., $\mathbb{P}(Z=1 \mid \sh = s) = s, \forall s \in [0,1]$, with $Z=1$ denoting an accurate detection in the sense that $\iou(\bh,B^{*}) \geq \beta$.
Similarly, by using an identity function for $\psi$, we obtain the definition proposed in \cite[Equation (3)]{oksuz2023towards}, i.e., $\mathbb{P}(Z= \iou(\bh,B^{*}) \mid \sh = s) = s, \forall s \in [0,1]$. Note that in both cases the requirement that the class prediction matches the ground truth class is simplified since we're considering a binary detector and all predictions share the same class as the ground truth.

\subsection{Measuring calibration}
A predictive function trained with risk minimization on a continuous domain will be exactly calibrated with probability 0.  We therefore need to measure a notion of calibration error. 
Analogous to the definition of calibration error for classification \cite{guo2017calibration}, we define the \textit{$L_1$ calibration error for a binary object detector $g_k$} for the $k$th class as:
\begin{align}\label{eq:LPcalibrationErrorOD}
    \operatorname{CE}(g_k) = \mathbb{E}\left[ \left| \mathbb{E} [\zh \mid \sh = s] - s \right| \right].
\end{align}

\paragraph{Estimating calibration error}
In practice, we rely on finite samples to estimate the quantities of interest. One of the desired properties of such estimators is \textbf{consistency}, i.e., the estimator $\widehat{\operatorname{CE}}_(g_k)$ should converge in probability to the true value $\operatorname{CE}(g_k)$~\cite{ibragimov2013statistical}: $\underset{n\to\infty}{\operatorname{plim}} \widehat{\operatorname{CE}}_(g_k) = \operatorname{CE}(g_k).$
Another desired property for calibration error estimators is \textbf{differentiability}, as it allows the estimator to be directly optimized alongside the task-specific loss function. Before presenting our estimator, which possesses both properties, we will introduce existing estimators and discuss their shortcomings.

In classification, one of the most widely used estimators of calibration error is the $L_1$ binned estimator, commonly referred to as ECE \cite{guo2017calibration}. 
Recent studies have proposed extensions of this estimator to the detection task, called detection ECE ($\dece$) \cite{kueppers2020} and Localization-aware ECE \cite{oksuz2023towards} ($\laece$), to be used as finite sample estimators of the notion of calibration as defined by a threshold link and an identity link, respectively.
Analogous to ECE, both $\dece$ and $\laece$ partition the predictions into $M$ equally-spaced bins. 
$\dece$ is computed by taking a weighted average of the difference between precision and confidence in each bin:
\begin{equation}
    \label{eq:dece_formulation}
    \dece = \sum_{m=1}^{M} \frac{|\mathcal{D}_m|}{|\mathcal{D}|}\left|\mathrm{prec}(m) - \mathrm{conf}(m)\right|,
\end{equation}
where $\mathcal{D}_m$ is the set of detections in $m^{th}$ bin, $|\mathcal{D}|$ is the total number of detections, $\mathrm{prec}(m)$ denotes the precision and $\mathrm{conf}(m)$ the average confidence over samples in the $m^{th}$ bin.
$\laece$ is computed as an average of $k$ per-class calibration errors obtained as:
\begin{equation}
    \label{eq:laece_formulation}
    \laece^k = \sum_{m=1}^{M} \frac{|\mathcal{D}_m^k|}{|\mathcal{D}^k|}\left|\mathrm{prec^k}(m) \times \mathrm{IoU}^k(m) - \mathrm{conf^k}(m)\right|,
\end{equation}

Several works have discussed the numerous flaws of binning estimators \cite{vaicenavicius2019,widmann2019,ding2020,ashukha2021}, such as sensitivity to the binning scheme, asymptotic inconsistency in many cases, and curse of dimensionality.
Moreover, these estimators are not differentiable and require additional approximations to be integrated as part of the optimization process.

As a remedy, \citet{zhang2020mix, popordanoska2022} recently proposed kernel density-based \cite{silverman86} estimators of calibration error, which are consistent, differentiable, and have more favorable statistical and computational properties. 
Therefore, we extend the estimator of \cite{popordanoska2022} to the detection task. That is, we wish to find an estimator:
\begin{align}
    \label{eq:calibration_error_estimator_od}
    \widehat{\operatorname{CE}(g_k)} = \frac{1}{w}\sum_{v=1}^w \left| \widehat{\mathbb{E}[Z \mid \sh = s_v]}-s_v \right|  ,
\end{align}
where $w$ denotes the number of detections across images.

We will focus on deriving an estimator of the conditional expectation, depending on the definition of calibration resulting from using two distinct link functions $\psi$ in Equation~\eqref{eq:binary_calibration_od}.
In case $\psi$ is a threshold function, Z is a discrete random variable, and the estimator obtained from this link function is identical to the one proposed in \cite{popordanoska2022}. 
However, in case $\psi$ is a continuous function such as the identity, Z is a continuous random variable, for which we derive the estimator below. 

Let $Z$ and $\sh$ be continuous random variables with joint density $f_{Z, \sh}(z,\hat{s})$. 
We wish to find an estimator for the conditional expectation:
\begin{align}
    \mathbb{E}[Z \mid \sh = s] &= \int_0^1 z \, p_{Z|\sh}(z \mid s) \, dz \\
    &= \frac{1}{p_{\sh}(s)} \int_0^1{z \, p_{Z, \sh}(z, s) \, dz} .
    \label{eq:estimator_EZS}
\end{align}
We derive an estimator for $p_{Z, \sh}(z, s)$ using KDE:
\begin{align}
    p_{Z, \sh}(z, s) &\approx \frac{1}{w} \sum_{u=1}^w \underbrace{k_s(\sh, s_u) k_{z}(Z, z_u)}_{k(Z, \sh, z_u, s_u)}.
\end{align}
where $k_s$ and $k_z$ denote any consistent kernels over their respective domains.  The resulting density estimate remains consistent \cite[Theorem~6]{Wied2010Consistency}.
Therefore, for the integral in Equation \eqref{eq:estimator_EZS}, we have:
\begin{align}
    \int_0^1{z \, p_{Z, \sh}(z, s) \, dz}
    &\approx \int_0^1 z \, \frac{1}{w} \sum_{u=1}^w k_s(\sh, s_u) k_z(Z, z_u) \, dz \\ 
    &= \frac{1}{w} \sum_{u=1}^w k_s(\sh, s_u) \int_0^1 z \,  k_{z}(Z, z_u) dz .
\end{align}

The final integral $\int_0^1 z \,  k_{z}(Z, z_u) dz$ converges to $z_u$ as the bandwidth of $k_z$ approaches zero, the density estimate approaches the true density, and the kernel $k_z$ approaches a Dirac delta function centered at $z_u$. 
Finally, the resulting estimator can be written as:
\begin{align}
    \widehat{\mathbb{E}[Z\mid \sh]} = \frac{\sum_{u=1}^w k(\sh, s_u) \, z_u}{\sum_{u=1}^w k(\sh, s_u)} .
\end{align}
Plugging this back into Equation~\eqref{eq:calibration_error_estimator_od}, and generalizing $z_u$ with a link function, we obtain the following \textit{estimator of $L_1$ calibration error for a binary object detector $g_k$}, a given link function $\psi$ and a similarity measure $\s$:
\begin{align}
    \label{eq:calibration_error_estimator_complete_form}
    \widehat{\operatorname{CE}(g_k)} = \frac{1}{w}\sum_{v=1}^w \left| \frac{\sum_{u \neq v} k(s_v, s_u) \, \psi(\s(b_u, b_u^{*}))}{\sum_{u \neq v} k(s_v, s_u)} - s_v \right| .
\end{align}
The estimator has values $\in [0, 1]$, it is consistent, asymptotically unbiased \cite{popordanoska2022}, and differentiable almost everywhere, as desired.

\subsection{Train-time calibration}
A common approach to designing calibration methods is to include a calibration-related auxiliary loss to the task-specific loss to be minimized during training \cite{munir2022towards, munir2023bridging, pathiraja2023multiclass}. In this section, we will show that existing auxiliary losses are not consistent estimators of calibration error. Instead, we propose to use our estimator derived in the previous section. 

\citet{munir2022towards} proposed an auxiliary loss $\operatorname{TCD} = \frac{1}{2}(d_{cls} + d_{det})$.
Adapting the notation from \cite[Equations (8) and (9)]{munir2022towards}, we have that $d_{cls} =  \mathbb{E}[| \sh - Y |]$ and $d_{det} = \mathbb{E}[|\iou - \sh|]$. We will derive a relationship between the classification term $d_{cls}$  and the classification calibration error, as well as between the localization term $d_{det}$ and the detection calibration error. 
The calibration error for binary classification is given by $\operatorname{CE}_{cls} := \mathbb{E}[|\mathbb{E}[Y=1|\sh] - \sh|]$, whereas $\operatorname{CE}_{det} := \mathbb{E}[|\mathbb{E}[IoU|\sh] - \sh|]$ defines the calibration error for detection when $\psi$ is an identity function. 

\begin{proposition}
    \label{prop:dcls}
    The classification calibration error upper bounds $d_{cls}$: 
    $\operatorname{CE}_{cls} \geq d_{cls}$. 
\end{proposition}

\begin{proposition}
    \label{prop:ddet}
    The detection calibration error is upper bounded by $ d_{det}$:
    $\operatorname{CE}_{det} \leq d_{det}$.
\end{proposition}
The proofs are given in Appendix. Thus, through Jensen's inequality, we showed that neither of these terms is a consistent estimator for any established concept of calibration error. We note that the auxiliary loss components defined in 
\cite[Equations (8) and (9)]{pathiraja2023multiclass} have the same functional form as $d_{cls}$ and $d_{det}$. 
BPC \cite[Equation (9)]{munir2023bridging} is based on a heuristic to maximize the confidence scores for accurate predictions and minimize scores for inaccurate ones. Therefore, none of the existing auxiliary losses are based on a consistent estimator of calibration error. Consequently, a principled way to minimize calibration error during training is to integrate our KDE-based estimator $\kde$, as the only consistent and differentiable estimator of calibration error for object detection, with the task specific-loss $\mathcal{L}_{det}$ as: 
\begin{align}
    \label{eq:object_detection_loss_kde}
    \mathcal{L} = \mathcal{L}_{det} + \lambda \kde, 
\end{align}
where $\lambda$ is a regularization parameter set by cross-validation.

\section{Experiments}

In this section, we investigate the performance of our estimator, both as a metric to evaluate calibration error, and as an auxiliary loss function used for train-time calibration.

\subsection{Setup}
\paragraph{Datasets}
We use the following three datasets:
\begin{enumerate}[nosep]
    \item MS COCO \cite{lin2014_coco} is a widely used object detection dataset that consists of more than 330000 object instances belonging to 80 object categories. It contains 118K training images and 5K validation images. 
    \item Cityscapes \cite{cordts2016_cityscapes} is an urban driving scene dataset, consisting of 2975 training and 500 validation images split into 8 object categories. 
    \item PASCAL VOC \cite{everingham2012_pascalvoc} has 20 classes, which are a subset of those in MS COCO. Following other works, for training we combine the VOC 2007 and VOC 2012 \texttt{trainval} sets, resulting in a total of 16551 images and more than 40000 objects. 
\end{enumerate}

We create separate validation sets with three different seeds, by spliting the original train set into new train and validation sets with 90:10 ratio. As the labels for the test set of COCO and Cityscapes are not publicly available, the original \texttt{val} set is used for reporting the results, and for VOC we evaluate on the VOC 2007 \texttt{test} set.

\paragraph{Object detectors}
We use three popular object detectors with different architectures, which have achieved strong performance in this task:
\begin{enumerate}[nosep]
    \item Faster R-CNN (F-RCNN) \cite{ren2015_fasterrcnn} is a popular two-stage detector with softmax classifier. The two-stage approach involves first generating a sparse set of object proposals, then refining them to accurately locate and classify the objects in the image. Typically they perform better compared to one-stage detectors, but they are slower and require more computational resources. 
    \item RetinaNet \cite{lin2017_retinanet} is a one-stage detector with sigmoid classifiers. Using this approach, objects are detected with a single pass through the network. RetinaNet applies a set of predefined anchor boxes to each feature level and predicts the objectness score and class probabilities for each anchor. In spite of their higher computation efficiency, one-stage detectors may have lower accuracy and struggle with small objects. 
    \item FCOS \cite{tian2019_fcos} is another common one-stage detector with sigmoid classifiers. Compared to RetinaNet, FCOS predicts the object bounding boxes and their class probabilities without requiring an anchor-based mechanism.
    Both FCOS and RetinaNet use focal loss, which aims to address the class imbalance problem, and it is also known to help with calibration. 
\end{enumerate}

\paragraph{Baselines}
Every object detector we consider has a combination of classification and regression loss, which we refer to as task-specific loss, and represents the first baseline we compare to.
Due to its strong performance, we include \textbf{TS} as a representative of the post-hoc calibration methods.
As a train-time baseline method, we compare with the recently proposed auxiliary loss function \textbf{TCD} \cite{munir2022towards}, which is added to the task-specific loss. 

\paragraph{Metrics}
For reporting detection performance we use the COCO-defined metrics. Namely, average precision AP denotes an average across categories and 10 (.50:.05:.95) $\iou$ overlap thresholds used for matching. In some experiments, we also report AP@0.5, AP@0.75, and AP for small, medium and large objects, as defined in the COCO challenge \cite{lin2014_coco}.
For measuring calibration error, we use our KDE-based estimator $\kde$ (Equation~\eqref{eq:calibration_error_estimator_complete_form}) with different link functions throughout the experiments. We use the same breakdown across object sizes and  $\iou$  overlaps as the one used for AP.
In addition, we report the $\dece$ metric (Equation~\eqref{eq:dece_formulation}) used in \cite{munir2022towards, munir2023bridging,pathiraja2023multiclass}, both at  $\iou$@0.5 and averaged over 10 thresholds (.50:.05:.95). In our synthetic experiment, we also compare with $\laece$ \cite{oksuz2023towards}. 

\paragraph{Implementation}

Traditionally, detections are obtained in two steps: generating predictions and post-processing. The latter step performs 
further processing to filter out redundant or low-confidence detections, through procedures like Non-Maximum Suppression (NMS) and top-$k$ selection, where typically $k=100$ for COCO. Choosing the optimal value for $k$, or a threshold $\gamma$ below which detections are rejected is an open area of research \cite{oksuz2018localization, oksuz2023towards}. For simplicity, in our experiments we set the score threshold $\gamma=0.5$ for our main results, and show an ablation study of the effect of this parameter on the reported metrics.  
Our implementation for evaluating CE on a test set is based on the official API for assessing detection performance on COCO, i.e.\, we utilize the same information about matching predicted and ground truth boxes and $\iou$ overlaps as for evaluating AP. 
We use a Beta kernel \cite{chen1999Beta} and set the bandwidth parameter with a leave-one-out maximum likelihood procedure.
For $\dece$ we adopt the implementation from \citet{kueppers2020}.
We rely on Detectron2 \cite{wu2019detectron2} for the implementation and training configuration of the detectors.
The code and trained models are available at \url{https://github.com/tpopordanoska/calibration-object-detection}.

\subsection{Results}

Here we empirically showcase the utilities of our proposed estimator and compare it with related work. 
The main purpose of an estimator of calibration error is to assess the extent to which a notion of calibration is violated. Therefore, we start our experiments by comparing the different variants of our KDE-based estimator (obtained with different choices for the link function) with the binning-based estimators $\dece$ and $\laece$. Subsequently, we provide a comprehensive evaluation of $\kde$ and the standard AP metric, at different $\iou$ thresholds and object sizes.
Finally, we demonstrate the differentiability of our estimator by incorporating it as an auxiliary loss function during training, both from scratch and for fine-tuning. We note that the estimator could also be integrated as part of a post-hoc method.

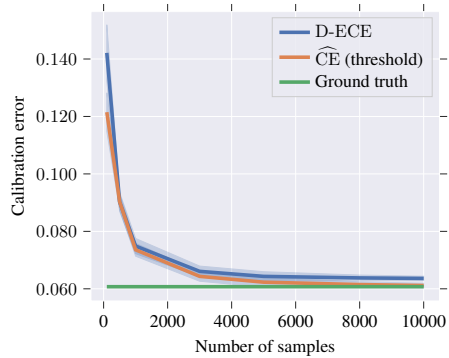
\begin{figure}[h]
    \centering
    \resizebox{0.35\textwidth}{!}{
\begin{tikzpicture}

\definecolor{color0}{rgb}{0.917647058823529,0.917647058823529,0.949019607843137}
\definecolor{color1}{rgb}{0.298039215686275,0.447058823529412,0.690196078431373}
\definecolor{color2}{rgb}{0.866666666666667,0.517647058823529,0.32156862745098}
\definecolor{color3}{rgb}{0.333333333333333,0.658823529411765,0.407843137254902}

\begin{axis}[
/pgf/number format/.cd, 1000 sep={},
axis background/.style={fill=color0},
axis line style={white},
legend cell align={left},
legend style={fill opacity=0.8, draw opacity=1, text opacity=1, draw=white!80!black, fill=color0},
scaled x ticks = false,
scaled y ticks = false,
tick align=outside,
x grid style={white},
x tick label style={/pgf/number format/fixed},
xlabel={Number of samples},
xmajorgrids,
xmin=-395, xmax=10495,
xtick style={color=white!15!black},
xtick={-2000,0,2000,4000,6000,8000,10000,12000},
xticklabels={\ensuremath{-}2000,0,2000,4000,6000,8000,10000,12000},
y grid style={white},
y tick label style={/pgf/number format/.cd, fixed, fixed zerofill, precision=3},
ylabel={Calibration error},
ymajorgrids,
ymin=0.0557755594515282, ymax=0.156504105702187,
ytick style={color=white!15!black}
]
\path [draw=color1, fill=color1, opacity=0.25]
(axis cs:100,0.151925535418066)
--(axis cs:100,0.132341464930883)
--(axis cs:500,0.0869903345344543)
--(axis cs:1000,0.0722347422504588)
--(axis cs:3000,0.0642452896097409)
--(axis cs:5000,0.0626967150030168)
--(axis cs:8000,0.06279734999873)
--(axis cs:10000,0.0628491466932229)
--(axis cs:10000,0.0643882096049625)
--(axis cs:10000,0.0643882096049625)
--(axis cs:8000,0.0647854729344074)
--(axis cs:5000,0.0659564567218122)
--(axis cs:3000,0.0679181131564493)
--(axis cs:1000,0.0775298913293411)
--(axis cs:500,0.0935401931983273)
--(axis cs:100,0.151925535418066)
--cycle;

\path [draw=color1, fill=color1, opacity=0.25]
(axis cs:100,0.128239547171723)
--(axis cs:100,0.11473152889954)
--(axis cs:500,0.0875630632159094)
--(axis cs:1000,0.0713526400834022)
--(axis cs:3000,0.0626374233595933)
--(axis cs:5000,0.0605594027087607)
--(axis cs:8000,0.0605347090312079)
--(axis cs:10000,0.0603541297356491)
--(axis cs:10000,0.0619530739744301)
--(axis cs:10000,0.0619530739744301)
--(axis cs:8000,0.0624156720093653)
--(axis cs:5000,0.0640605319991193)
--(axis cs:3000,0.0660293277271186)
--(axis cs:1000,0.0757884469718041)
--(axis cs:500,0.0931208833935877)
--(axis cs:100,0.128239547171723)
--cycle;

\addplot [line width=2pt, color1]
table {%
100 0.142133500174475
500 0.0902652638663908
1000 0.0748823167898999
3000 0.0660817013830951
5000 0.0643265858624145
8000 0.0637914114665687
10000 0.0636186781490927
};
\addlegendentry{$\dece$}
\addplot [line width=2pt, color2]
table {%
100 0.121485538035631
500 0.0903419733047485
1000 0.0735705435276031
3000 0.0643333755433559
5000 0.06230996735394
8000 0.0614751905202866
10000 0.0611536018550396
};
\addlegendentry{$\kde$ (threshold)}
\addplot [line width=2pt, color3]
table {%
100 0.0607329308986664
500 0.0607329308986664
1000 0.0607329308986664
3000 0.0607329308986664
5000 0.0607329308986664
8000 0.0607329308986664
10000 0.0607329308986664
};
\addlegendentry{Ground truth}
\end{axis}

\end{tikzpicture}
    }
    \caption{Comparison of $\dece$ vs. $\kde$ (threshold link) as a function of the number of points used for the estimation. }
    \label{fig:comparison_of_metrics}
\end{figure}

\paragraph{Comparison of metrics}
$\dece$ measures the notion of calibration obtained by letting $\psi$ be a threshold function, whereas $\laece$ is used for assessing calibration as obtained through an identity function. We compare these binning-based estimators to the corresponding KDE-based estimators on a synthetic binary problem, where the ground truth calibration error is known. Similar to \cite{popordanoska2022}, we apply temperature scaling with $t_1 = 0.6$ to a uniform sample of numbers in $[0, 1]$. The purpose of this step is to ensure that the confidence scores are concentrated around $0$ and $1$, as in a realistic scenario. Then, we sample the labels, denoting $\mathbbm{1}[\iou(\bh, B*) \geq \beta]$, according to that distribution, and therefore obtain perfect calibration. In order to simulate miscalibration, we perform another temperature scaling with $t_2 = 0.6$. 
$\dece$ is calculated using 20 equally-spaced bins, as in \cite{kueppers2020}. For $\laece$ we use 25 bins, following \cite{oksuz2023towards}, and compute the weighted difference between $t_1$ and $t_2$ scaled scores in each bin.
As shown in Figure~\ref{fig:comparison_of_metrics}, $\dece$ and $\kde$ have similar performance and both yield good CE estimates with a few thousand points. The comparison with $\laece$ can be found in Appendix.

\paragraph{Evaluating a model zoo}
We conduct a rigorous analysis to determine the calibration of F-RCNN, RetinaNet and FCOS on popular benchmark datasets like COCO, Cityscapes and Pascal VOC. Analogous to the evaluation on the COCO challenge, we estimate calibration error across different sizes of objects, and for different thresholds for the $\iou$ overlap required to determine the matching between predicted and ground truth box. For this evaluation, we use a threshold link function. The results are summarized in Tables \ref{tab:map_model_zoo} and \ref{tab:ece_model_zoo}. The first thing to note is that modern object detectors are intrinsically miscalibrated. When comparing the different architectures, an observation can be made that RetinaNet and FCOS have lower calibration errors than F-RCNN. This may be attributed to the use of focal loss for the classification branch of RetinaNet and FCOS, which has been empirically shown to improve calibration. Moreover, different metrics reveal complementary information about the calibration of the detector. For instance, CE$_{50}$ on Pascal VOC shows similar performance of F-RCNN and FCOS, whereas CE$_{75}$ reveals that RetinaNet and FCOS have much better performance than F-RCNN for a calibration error that requires an $\iou$ overlap of 0.75 for a correct detection. Similarly, both RetinaNet and FCOS demonstrate similar calibration errors when detecting small objects. However, FCOS exhibits superior calibration scores for medium-sized objects, despite both detectors achieving the same AP$_{M}$.

\begin{table*}[ht]
    \centering
    \caption{Detection performance of object detectors on three popular datasets.}
    \resizebox{0.8\textwidth}{!}{
    \begin{tabular}{cccccccc}
        \toprule
        Dataset & Model & AP & AP$_{50}$ & AP$_{75}$ & AP$_{S}$ & AP$_{M}$ & AP$_{L}$ \\
         \midrule
          \multirow{3}{*}{COCO} & F-RCNN & $36.11_{\pm 0.10}$ & $53.35_{\pm 0.15}$ & $40.04_{\pm 0.12}$ & $18.88_{\pm 0.10}$ & $39.40_{\pm 0.11}$ &  $47.98_{\pm 0.08}$  \\
           & RetinaNet & $30.83_{\pm 0.12}$ & $43.33_{\pm 0.22}$ & $34.03_{\pm 0.07}$ & $13.59_{\pm 0.28}$ & $34.13_{\pm 0.13}$ &  $43.16_{\pm 0.24}$ \\
           & FCOS & $34.02_{\pm 0.06}$ & $48.84_{\pm 0.04}$ & $37.37_{\pm 0.10}$ & $17.29_{\pm 0.17}$ & $37.84_{\pm 0.17}$ & $45.15_{\pm 0.17}$ \\
        \midrule
          \multirow{3}{*}{Cityscapes} & F-RCNN & $35.36_{\pm 0.65}$ & $54.46_{\pm 0.82}$ & $37.90_{\pm 0.79}$ & $11.54_{\pm 0.31}$ & $35.34_{\pm 0.51}$ & $56.62_{\pm 0.48}$ \\
          & RetinaNet & $34.60_{\pm 0.23}$ & $52.60_{\pm 0.81}$ & $36.41_{\pm 0.35}$ & $10.39_{\pm 0.15}$ & $35.56_{\pm 0.15}$ &  $57.47_{\pm 0.39}$  \\
          & FCOS & $34.81_{\pm 0.08}$ & $52.31_{\pm 0.29}$ & $36.62_{\pm 0.23}$ & $10.66_{\pm 0.60}$ & $32.84_{\pm 0.44}$ & $56.90_{\pm 0.81}$ \\
        \midrule
          \multirow{3}{*}{Pascal VOC} & F-RCNN & $52.96_{\pm 0.05}$ & $75.88_{\pm 0.08}$ & $59.67_{\pm 0.05}$ & $18.48_{\pm 0.39}$ & $40.90_{\pm 0.11}$ & $61.43_{\pm 0.08}$ \\
          & RetinaNet & $53.20_{\pm 0.02}$ & $73.36_{\pm 0.02}$ & $58.75_{\pm 0.11}$ & $16.39_{\pm 0.42}$ & $40.19_{\pm 0.26}$ & $62.06_{\pm 0.02}$  \\
          & FCOS & $52.13_{\pm 0.02}$ & $73.50_{\pm 0.07}$ & $57.63_{\pm 0.04}$ & $19.23_{\pm 0.23}$ & $40.52_{\pm 0.12}$ & $60.28_{\pm 0.06}$ \\
        \bottomrule
    \end{tabular}
    }
    \label{tab:map_model_zoo}
\end{table*}

\begin{table*}[ht]
    \centering
    \caption{Calibration performance (CE $\times$ 100) of object detectors on three popular datasets. The calibration error is broken down into variants depending on the $\iou$ threshold and the size of the objects, same as for AP. }
    \resizebox{0.8\textwidth}{!}{
    \begin{tabular}{cccccccc}
        \toprule
        Dataset & Model & $\kde$ & $\kde_{50}$ & $\kde_{75}$ & $\kde_{S}$ & $\kde_{M}$ & $\kde_{L}$ \\
        \midrule
         \multirow{3}{*}{COCO} & F-RCNN & $37.33_{\pm 0.09}$ & $20.48_{\pm 0.10}$ & $32.56_{\pm 0.15}$ & $35.72_{\pm 0.51}$ & $38.94_{\pm 0.10}$ & $40.81_{\pm 0.12}$   \\
        & RetinaNet & $21.89_{\pm 0.36}$ & $12.03_{\pm 0.19}$ & $14.70_{\pm 0.41}$ & $26.73_{\pm 0.52}$ & $26.04_{\pm 0.13}$ & $27.58_{\pm 0.17}$ \\
        & FCOS & $24.40_{\pm 0.13}$ & $16.55_{\pm 0.21}$ & $19.78_{\pm 0.17}$ & $26.34_{\pm 0.39}$ & $26.33_{\pm 0.24}$ & $29.68_{\pm 0.03}$ \\
        \midrule
          \multirow{3}{*}{Cityscapes} & F-RCNN & $37.45_{\pm 0.46}$ & $17.00_{\pm 0.46}$ & $32.55_{\pm 0.48}$ & $40.85_{\pm 5.08}$ & $41.15_{\pm 1.57}$ & $38.22_{\pm 0.78}$ \\
          & RetinaNet & $32.22_{\pm 0.29}$ & $12.91_{\pm 0.95}$ & $28.38_{\pm 0.74}$ & $29.99_{\pm 0.57}$ & $38.16_{\pm 1.27}$ & $35.91_{\pm 0.46}$ \\
          & FCOS & $26.29_{\pm 0.98}$ & $13.91_{\pm 1.28}$ & $21.09_{\pm 1.57}$ & $28.44_{\pm 1.66}$ & $32.71_{\pm 0.78}$ & $28.13_{\pm 0.93}$ \\
        \midrule
          \multirow{3}{*}{Pascal VOC} & F-RCNN & $34.88_{\pm 0.10}$ & $17.03_{\pm 0.16}$ & $28.75_{\pm 0.17}$ & $47.67_{\pm 0.91}$ & $41.99_{\pm 0.33}$ & $30.84_{\pm 0.07}$ \\
          & RetinaNet & $24.11_{\pm 0.09}$ & $7.95_{\pm 0.19}$ & $17.89_{\pm 0.14}$ & $36.31_{\pm 1.22}$ & $31.53_{\pm 0.06}$ & $21.73_{\pm 0.11}$ \\
          & FCOS & $22.32_{\pm 0.05}$ & $15.77_{\pm 0.10}$ & $15.55_{\pm 0.11}$ & $36.84_{\pm 0.65}$ & $27.12_{\pm 0.04}$ & $20.97_{\pm 0.02}$ \\
        \bottomrule
    \end{tabular}
    }
    \label{tab:ece_model_zoo}
\end{table*}

\paragraph{Calibration-regularized training}
Another notion of calibration that might be of interest is when the link $\psi$ is an identity function. Intuitively, this type of calibration requires that the predicted score corresponds to the $\iou$ overlap with a ground truth box. Using this setup for $\kde$, we performed extensive experiments to compare the performance of our estimator with a post-hoc method (TS), and with a train-time loss (TCD). The temperature is chosen on a validation set by minimizing NLL. Table \ref{table:training_cityscapes} reports the AP, $\kde$, and $\dece$ with $\iou$ overlap of 0.5, for RetinaNet and FCOS, trained on Cityscapes. 

\begin{table}[ht]
	\centering
	\caption{Comparison of detection (AP) and calibration performance (CE $\times$ 100) of models trained with RetinaNet and FCOS on Cityscapes. Models with no calibration, with post-hoc (TS) and train-time (TCD) methods are compared with our approach. }
 	\resizebox{0.95\linewidth}{!}{%
		\begin{tabular}{cccc}
			\toprule
			Model &  AP $\uparrow$ & $\kde$ $\downarrow$ & $\dece_{50}$ $\downarrow$ \\
			\midrule
			RetinaNet & $34.60_{\pm 0.23}$ & $23.25_{\pm 0.39}$ & $12.54_{\pm 0.66}$ \\
			RetinaNet + TS & $34.60_{\pm 0.23}$ & $18.67_{\pm 2.07}$ & $11.05_{\pm 1.25}$ \\
                RetinaNet + TCD & $33.94_{\pm 0.81}$ & $25.01_{\pm 1.13}$ & $13.04_{\pm 0.47}$ \\
                \rowcolor{maroon!10}
			RetinaNet + $\kde$& $32.59_{\pm 0.71}$ & $17.33_{\pm 3.15}$  & $11.07_{\pm 0.22}$ \\
   			\midrule
			FCOS & $34.81_{\pm 0.08}$ & $14.43_{\pm 1.31}$ & $13.23_{\pm 1.18}$ \\
			FCOS + TS & $33.62_{\pm 0.44}$ & $13.33_{\pm 2.16}$ & $11.96_{\pm 0.90}$ \\
                FCOS + TCD & $35.57_{\pm 0.23}$ & $16.85_{\pm 1.27}$ & $13.68_{\pm 0.86}$ \\
                \rowcolor{maroon!10}
			FCOS + $\kde$& $33.73_{\pm 0.57}$ & $13.07_{\pm 0.83}$ & $17.45_{\pm 0.61}$ \\
			\bottomrule
		\end{tabular}%
  }
	\label{table:training_cityscapes}
\end{table}

\paragraph{Fine-tuning}
Since there already exist a variety of trained object detectors on COCO, we demonstrate that the benefits of adding our estimator as an auxiliary loss can be observed also for fine-tuning pre-trained models even for a few epochs. Table~\ref{table:finetuning_coco} shows the reduction in CE, achieved by fine-tuning for three epochs, by minimizing the loss defined in Equation~\eqref{eq:object_detection_loss_kde} on COCO dataset. 

\begin{table}[h]
\centering
	\caption{Performance (AP and CE $\times$ 100) of models without calibration and fine-tuned models using our auxiliary loss on COCO.}
  	\resizebox{0.7\linewidth}{!}{%
		\begin{tabular}{ccc}
			\toprule
			Model & AP $\uparrow$ & $\kde$ $\downarrow$ \\
			\midrule
			F-RCNN & $36.11_{\pm 0.10}$ &  $31.76_{\pm 0.05}$ \\
                \rowcolor{maroon!10}
                F-RCNN + $\kde$& $34.72_{\pm 0.09}$ & $26.91_{\pm 0.14}$ \\
			\midrule
			RetinaNet & $30.83_{\pm 0.12}$ & $21.89_{\pm 0.36}$ \\
                \rowcolor{maroon!10}
                RetinaNet + $\kde$& $30.10_{\pm 0.06}$ & $9.72_{\pm 0.18}$ \\
   			\midrule
			FCOS & $34.02_{\pm 0.06}$ & $24.40_{\pm 0.13}$ \\
                \rowcolor{maroon!10}
                FCOS + $\kde$ & $33.41_{\pm 0.12}$ & $15.32_{\pm 0.17}$ \\
			\bottomrule
		\end{tabular}%
  }
	\label{table:finetuning_coco}
\end{table}

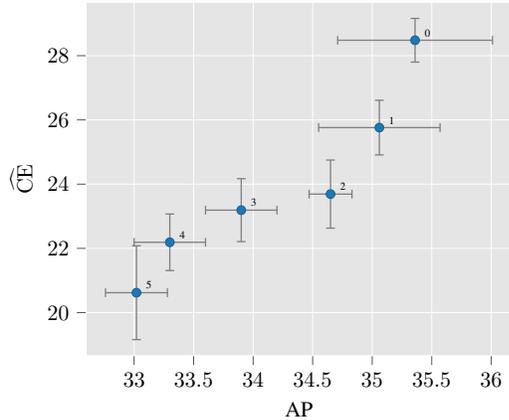
\begin{figure}[t]
    \centering
    \resizebox{0.4\textwidth}{!}{
\begin{tikzpicture}

\begin{axis}[
axis background/.style={fill=white!89.8039215686275!black},
axis line style={white},
scaled y ticks = false,
tick align=outside,
tick pos=left,
x grid style={white},
xlabel={AP},
xmajorgrids,
xmin=32.5975, xmax=36.1725,
xtick style={color=white!33.3333333333333!black},
y grid style={white},
y tick label style={/pgf/number format/fixed},
ylabel={$\kde$},
ymajorgrids,
ymin=18.66, ymax=29.66,
ytick style={color=white!33.3333333333333!black}
]
\path [draw=white!50.1960784313725!black, semithick]
(axis cs:34.71,28.48)
--(axis cs:36.01,28.48);

\path [draw=white!50.1960784313725!black, semithick]
(axis cs:34.55,25.76)
--(axis cs:35.57,25.76);

\path [draw=white!50.1960784313725!black, semithick]
(axis cs:34.47,23.69)
--(axis cs:34.83,23.69);

\path [draw=white!50.1960784313725!black, semithick]
(axis cs:33.6,23.19)
--(axis cs:34.2,23.19);

\path [draw=white!50.1960784313725!black, semithick]
(axis cs:33,22.19)
--(axis cs:33.6,22.19);

\path [draw=white!50.1960784313725!black, semithick]
(axis cs:32.76,20.62)
--(axis cs:33.28,20.62);

\path [draw=white!50.1960784313725!black, semithick]
(axis cs:35.36,27.8)
--(axis cs:35.36,29.16);

\path [draw=white!50.1960784313725!black, semithick]
(axis cs:35.06,24.91)
--(axis cs:35.06,26.61);

\path [draw=white!50.1960784313725!black, semithick]
(axis cs:34.65,22.63)
--(axis cs:34.65,24.75);

\path [draw=white!50.1960784313725!black, semithick]
(axis cs:33.9,22.21)
--(axis cs:33.9,24.17);

\path [draw=white!50.1960784313725!black, semithick]
(axis cs:33.3,21.31)
--(axis cs:33.3,23.07);

\path [draw=white!50.1960784313725!black, semithick]
(axis cs:33.02,19.16)
--(axis cs:33.02,22.08);

\addplot [semithick, white!50.1960784313725!black, mark=|, mark size=2, mark options={solid}, only marks]
table {%
34.71 28.48
34.55 25.76
34.47 23.69
33.6 23.19
33 22.19
32.76 20.62
};
\addplot [semithick, white!50.1960784313725!black, mark=|, mark size=2, mark options={solid}, only marks]
table {%
36.01 28.48
35.57 25.76
34.83 23.69
34.2 23.19
33.6 22.19
33.28 20.62
};
\addplot [semithick, white!50.1960784313725!black, mark=-, mark size=2, mark options={solid}, only marks]
table {%
35.36 27.8
35.06 24.91
34.65 22.63
33.9 22.21
33.3 21.31
33.02 19.16
};
\addplot [semithick, white!50.1960784313725!black, mark=-, mark size=2, mark options={solid}, only marks]
table {%
35.36 29.16
35.06 26.61
34.65 24.75
33.9 24.17
33.3 23.07
33.02 22.08
};
\draw (axis cs:35.41,28.4799) node[
  scale=0.5,
  anchor=south west,
  text=black,
  rotate=0.0
]{0};
\draw (axis cs:35.11,25.7599) node[
  scale=0.5,
  anchor=south west,
  text=black,
  rotate=0.0
]{1};
\draw (axis cs:34.7,23.6899) node[
  scale=0.5,
  anchor=south west,
  text=black,
  rotate=0.0
]{2};
\draw (axis cs:33.95,23.1899) node[
  scale=0.5,
  anchor=south west,
  text=black,
  rotate=0.0
]{3};
\draw (axis cs:33.35,22.1899) node[
  scale=0.5,
  anchor=south west,
  text=black,
  rotate=0.0
]{4};
\draw (axis cs:33.07,20.6199) node[
  scale=0.5,
  anchor=south west,
  text=black,
  rotate=0.0
]{5};
\addplot [
  colormap={mymap}{[1pt]
 rgb(0pt)=(0.12156862745098,0.466666666666667,0.705882352941177);
  rgb(1pt)=(1,0.498039215686275,0.0549019607843137)
},
  only marks,
  scatter,
  scatter src=explicit
]
table [x=x, y=y, meta=colordata]{%
x  y  colordata
35.36 28.48 1.0
35.06 25.76 1.0
34.65 23.69 1.0
33.9 23.19 1.0
33.3 22.19 1.0
33.02 20.62 1.0
};
\end{axis}

\end{tikzpicture}
    }
    \caption{Effect of $\lambda$ on $\operatorname{AP}$ and $\kde$. The points represent fine-tuned Faster-RCNN detectors on Cityscapes for three epochs. The number next to the point denotes the value of $\lambda$.}
    \label{fig:lambda_cityscapes_finetuning}
\end{figure}

\paragraph{Ablation study}
We investigate the effect of the regularization parameter $\lambda$, the test score threshold $\gamma$, and the number of epochs used for fine-tuning on the reported metrics. 
In Figure~\ref{fig:lambda_cityscapes_finetuning} we present fine-tuned F-RCNN detectors on Cityscapes for different values of the $\lambda$ parameter. We notice that increasing the weight of calibration regularization leads to noticeable reduction in calibration error. 
Table~\ref{tab:test_score_threshold_cityscapes_finetuning} shows the evaluated performance on test set using different thresholds $\gamma$. From Table~\ref{tab:num_epochs_cityscapes_finetuning} we can observe that fine-tuning even for a few epochs can lead to improvements in calibration error, without sacrificing AP.
Further experiments for the effect of $\lambda$, both for fine-tuned models and trained from scratch on the three datasets, can be found in the Appendix.

\begin{table}[ht]
    \centering
    \caption{ Effect of test score threshold $\gamma$ on detection performance and calibration of Faster-RCNN trained on Cityscapes.}
    \resizebox{0.35\textwidth}{!}{
    \begin{tabular}{ccc}
        \toprule
        Model & AP & $\kde$ \\
        \midrule
        F-RCNN ($\gamma$ = 0.1) & $38.34_{\pm 0.18}$ & $22.61_{\pm 1.11}$ \\ 
        F-RCNN ($\gamma$ = 0.2) & $37.53_{\pm 0.26}$ & $26.26_{\pm 0.78}$ \\
        F-RCNN ($\gamma$ = 0.3) & $36.92_{\pm 0.31}$ & $27.86_{\pm 0.80}$ \\
        F-RCNN ($\gamma$ = 0.4) & $36.15_{\pm 0.43}$ & $28.36_{\pm 0.83}$ \\
        F-RCNN ($\gamma$ = 0.5) & $35.36_{\pm 0.65}$ & $28.48_{\pm 0.68}$ \\
        F-RCNN ($\gamma$ = 0.6) & $34.31_{\pm 0.75}$ & $28.19_{\pm 0.90}$ \\
        F-RCNN ($\gamma$ = 0.7) & $33.26_{\pm 0.96}$ & $27.48_{\pm 0.99}$ \\
         \bottomrule
    \end{tabular}
    }
    \label{tab:test_score_threshold_cityscapes_finetuning}
\end{table}

\begin{table}[h]
    \centering
    \caption{Effect of number of epochs for fine-tuning a Faster-RCNN network on Cityscapes, with $\lambda = 1$ and $\gamma= 0.5$ .}
    \resizebox{0.35\textwidth}{!}{
    \begin{tabular}{ccc}
        \toprule
        Model & AP & $\kde$ \\
        \midrule
        F-RCNN & $35.36_{\pm 0.65}$ & $28.48_{\pm 0.68}$ \\
        F-RCNN + $\kde$ (1x) & $34.39_{\pm 0.19}$ & $25.88_{\pm 1.40}$ \\
        F-RCNN + $\kde$ (2x) & $34.45_{\pm 0.22}$ & $25.48_{\pm 0.60}$ \\
        F-RCNN + $\kde$ (3x) & $35.06_{\pm 0.51}$ & $25.76_{\pm 0.85}$ \\
        F-RCNN + $\kde$ (4x) & $35.67_{\pm 0.41}$ & $26.72_{\pm 0.56}$ \\
        F-RCNN + $\kde$ (5x) & $35.80_{\pm 0.40}$ & $26.19_{\pm 1.46}$ \\
         \bottomrule
    \end{tabular}
    }
    \label{tab:num_epochs_cityscapes_finetuning}
\end{table}

\section{Discussion}

In this paper, we tackled the challenge of defining calibration and estimating calibration error for object detection. 
Specifically, calibration error for OD is a quantity that depends on the choice of a measure of ``correctness`` of a prediction, which we have decomposed into a similarity measure $L$ and a link function $\psi$, which together determine the notion of calibration that is relevant for a particular scenario. Beyond the definition, we also proposed a consitent and differentiable estimator of a calibration error for OD, which can be used for common one-stage, two-stage, anchor-based and anchor-free DNN-based detectors.

The empirical results showed that our estimator performs equally well for estimating calibration error as existing binning-based versions, while offering superior statistical properties. Moreover, due to its differentiability, it can be directly included as part of both post-hoc and train-time calibration setups. In particular, when integrated into the task-specific loss during training, our estimator achieves substantial improvements in calibration error, while maintaining similar levels of AP. We have also shown that a reduction in calibration error can be achieved by fine-tuning with our auxiliary loss for a few epochs, thus eliminating the need to train object detectors from scratch. 

This work also has some limitations. The main limitation is the intrinsic $\mathcal{O}(n^2)$~\cite{popordanoska2022} complexity of KDE. When using the estimator of calibration error as an auxiliary loss for one-stage object detectors, which generate a considerable number of candidate detections during training, the complexity becomes particularly challenging. Nevertheless, our synthetic experiments have revealed that the calibration error can be accurately estimated using only a few thousand data points. As a result, during training with RetinaNet and FCOS, we opt to randomly sample a subset of detections and evaluate the calibration error based on this reduced set of data points.
Our empirical results verify that this is an effective strategy to conduct calibration-regularized training.
Another limitation of this paper is that we did not include a transformer-based model, such as DINO~\cite{li2023mask}, in our experiments. However, the presented mathematical principles are model-agnostic and our estimator can be trivially extended to this family of detectors as well. 
Finally, it is worth noting that in certain scenarios, calibration regularized training results in a slight reduction of the AP metric. Exploring the relationship between CE, AP, and the test score threshold, along with investigating the advantage that calibration brings to downstream tasks that utilize object detectors, is an interesting direction for future work.

To conclude, our study puts forward a \textit{novel and principled} view on calibration in OD, and proposes a new mathematical framework, which enables both estimation of calibration error, and calibration regularized training for object detection. Considering the critical role of calibration in enhancing overall system robustness, our method is highly relevant across diverse computer vision applications, and especially in risk-sensitive scenarios.

\section*{Acknowledgements}

This research received funding from the Flemish Government (AI Research Program) and the Research Foundation - Flanders (FWO) through project number G0G2921N. The publication was also supported by funding from the
Academy of Finland (Profi6 336449 funding program). The authors wish to acknowledge CSC -- IT Center for Science, Finland, for generous computational resources.

{\small
\bibliographystyle{plainnat}
\bibliography{egbib}
}

\end{document}